# On Partially Controlled Multi-Agent Systems

**Ronen I. Brafman** BRAFMAN@CS.UBC.CA
*Computer Science Department*
*University of British Columbia*
*Vancouver, B.C., Canada V6L 1Z4*

**Moshe Tennenholtz** MOSHET@IE.TECHNION.AC.IL
*Industrial Engineering and Management*
*Technion - Israel Institute of Technology*
*Haifa 32000, Israel*

## Abstract

Motivated by the control theoretic distinction between *controllable* and *uncontrollable* events, we distinguish between two types of agents within a multi-agent system: *controllable agents*, which are directly controlled by the system's designer, and *uncontrollable agents*, which are not under the designer's direct control. We refer to such systems as *partially controlled multi-agent systems*, and we investigate how one might influence the behavior of the uncontrolled agents through appropriate design of the controlled agents. In particular, we wish to understand which problems are naturally described in these terms, what methods can be applied to influence the uncontrollable agents, the effectiveness of such methods, and whether similar methods work across different domains. Using a game-theoretic framework, this paper studies the design of partially controlled multi-agent systems in two contexts: in one context, the uncontrollable agents are expected utility maximizers, while in the other they are reinforcement learners. We suggest different techniques for controlling agents' behavior in each domain, assess their success, and examine their relationship.

## 1. Introduction

The control of agents is a central research topic in two engineering fields: Artificial Intelligence (AI) and Discrete Events Systems (DES) (Ramadge & Wonham, 1989). One particular area both of these fields have been concerned with is multi-agent environments; examples include work in *distributed AI* (Bond & Gasser, 1988), and work on *decentralized supervisory control* (Lin & Wonham, 1988). Each of these fields has developed its own techniques and has incorporated particular assumptions into its models. Hence, it is only natural that techniques and assumptions used by one field may be adopted by the other or may lead to new insights for the other field.

In difference to most AI work on multi-agent systems, work on decentralized discrete event systems distinguishes between *controllable* and *uncontrollable* events. Controllable events are events that can be directly controlled by the system's designer, while uncontrollable events are not directly controlled by the system's designer. Translating this terminology into the context of multi-agent systems, we introduce the distinction between two types of agents: *controllable agents*, which are directly controlled by the system's designer, and *uncontrollable agents*, which are not under the designer's direct control. This leads naturally to the concept of *partially controlled multi-agent system* (PCMAS) and to the following design challenge: ensuring that *all* agents in the system behave appropriately through adequate design of the controllable agents. We believe that many problems are naturally formulated as instances of PCMAS design. Our goal is to characterize important instances of this design problem, to examine the tools that can be used to solve it, and to assess the effectiveness and generality of these tools.

What distinguishes partially controlled multi-agent systems in the AI context from similar models in DES are the structural assumptions we make about the uncontrolled agents involved. Unlike typical DES models which are concerned with physical processes or devices, AI is particularly interested in self-motivated agents, two concrete examples of which are rational agents, i.e., expected utility maximizers, and learning agents, e.g., reinforcement learners. Indeed, these examples constitute the two central models of self-motivated agents in game theory and decision theory, referred to as the *educative* and *evolutive* models (e.g., see Gilboa & Matsui, 1991). The special nature of the uncontrollable agents and the special structure of the uncontrollable events they induce is what differentiates PCMAS from corresponding models in the DES literature. This difference raises new questions and suggests a new perspective on the design of multi-agent systems. In particular, it calls for techniques for designing controllable agents that, by exploiting the structural assumptions, can influence the behavior of the uncontrollable agents and lead the system to a desired behavior.

In order to understand these issues, we study two problems that can be stated and solved by adopting the perspective of PCMAS design; problems which by themselves should be of interest to a large community. In both of these problems our goal is to influence the behavior of agents that are not under our control. We exert this influence indirectly by choosing suitable behaviors for those agents that are under our direct control. In one case, we attempt to influence the behavior of rational agents, while in the other case, we try to influence learning agents.

Our first study is concerned with the enforcement of social laws. When a number of agents designed by different designers work within a shared environment, it can be beneficial to impose certain constraints on their behavior, so that, overall, the system will function better. For example, Shoham and Tennenholtz (1995) show that by imposing certain "traffic laws," they can considerably simplify the task of motion planning for each robot, while still enabling efficient motions. Indeed, as we see later, such conventions are at the heart of many coordination techniques in multi-agent systems. Yet, without suitable mechanisms, rational agents may have an incentive not to follow these conventions. We show how, in certain cases, we can use the perspective of partially controlled multi-agent systems and the structural assumption of rationality to enforce these conventions.

Our second study involves a two-agent system consisting of a teacher and a student. The teacher is a knowledgeable agent, while the student is an agent that is learning how to behave in its domain. Our goal is to utilize the teacher (which is under our control) to improve the behavior of the student (which is not controlled by us). Hence, this is an instance of partially controlled multi-agent systems in which the structural assumption is that the uncontrolled agent employs a particular learning algorithm.

Both studies presented in this paper suggest techniques for achieving satisfactory system behavior through the design of the controllable agents, and where relevant, these techniques are experimentally assessed. Beyond the formulation and solution of two interesting problems in multi-agent system design, this paper suggests a more general perspective on certain design problems. Although we feel that it is still premature to draw general conclusion about the potential for a general theory of PCMAS design, certain concepts, those of punishment and reward, suggest themselves as central to this area.

The paper is organized as follows: In Section 2, we describe the problem of enforcing social behavior in multi-agent systems. In Section 3 we describe a standard game-theoretic model for this problem and suggest the mechanism of threats and punishments as a general tool for this class of problems. Issues that pertain to the design of threats and punishments are discussed in Section 4. Section 5 introduces our second case study in PCMAS design: embedded teaching of reinforcement learners. In this context, a teacher and a learner are embedded in a shared environment with the teacher serving as the controller whose aim is to direct the learner to a desired behavior. A formal model of this problem is introduced in Section 6. In Section 7, we show how to derive optimal teaching policies (under certain assumptions) by viewing teaching as a Markov decision process. The effectiveness of different teaching policies is studied experimentally in Section 8. Finally, in Section 9, we examine the relationship between the methods used in each of the two domains and the possibility of a general methodology for designing partially controlled multi-agent systems. We conclude in Section 10, with a summary and discussion of related work.

## 2. The Enforcement of Social Behavior

In this section we introduce the problem of the enforcement of social laws in a multi-agent context. Our proposed solution falls naturally out of the PCMAS design perspective we take. Here, we explain and motivate the particular problem of social law enforcement and our approach to its solution. In Sections 3 and 4 we formalize and investigate this approach in the framework of a general game-theoretic model.

We use the following scenario to illustrate the problem:

> You have been hired to design a new working environment for artificial agents. Part of your job involves designing a number of agents that will use and maintain a warehouse. Other agents, designed by different designers, will be using the warehouse to obtain equipment. To make sure that different agents designed by different designers can operate efficiently in this environment, you choose to introduce a number of social laws, that is, constraints on the behavior of agents, that will help the agents coordinate their activities in this domain. These rules include a number of 'traffic laws', regulating motion in the domain, as well as a law that specifies that every tool that is used by an agent must be returned to its designated storage area. Your robots are programmed to follow these laws, and you expect the others to do so. Your laws are quite successful, and allow efficient activity in the warehouse, until a new designer arrives. Pressed by his corporate bosses to deliver better performance, he decides to exploit all your rules. He designs his agent to locally maximize its performance, regardless of the social laws. What can you do?

In multi-participant environments, as the one above, each agent might have its own dynamic goals, and we are interested in finding ways in which agents can coexist while achieving their goals. Several approaches for coordination of agent activity are discussed in the distributed systems and the DAI literature. Some examples are: protocols for reaching consensus (Dwork & Moses, 1990), rational deals and negotiations (Zlotkin & Rosenschein, 1993; Kraus & Wilkenfeld, 1991; Rosenschein & Genesereth, 1985), organizational structures (Durfee, Lesser, & Corkill, 1987; Fox, 1981; Malone, 1987), and social laws (Moses & Tennenholtz, 1995; Shoham & Tennenholtz, 1995; Minsky, 1991; Briggs & Cook, 1995). In some of these methods, the behavior of an agent is predetermined or prescribed from a certain stage, for example, the content of the deal after it is reached, the outcome of the negotiation process after it is completed, or the social law after it is instituted. This work relies on the assumption that the agents *follow* these prescribed behaviors, e.g., they obey the law or stick to the agreement. This assumption is central to the success of any of these methods. However, it makes agents that follow the rules vulnerable to any rational agent that performs local maximization of payoff, exploiting the knowledge that others follow the rules. In our example, the new designer may program his robot not to return the tools, saving the time required to do so, thus causing other agents to fail in their tasks.

Despite its somewhat futuristic flavor (although instances of such shared environments are beginning to appear in cyberspace), this scenario is useful in illustrating the vulnerability of some of the most popular coordination mechanism appearing in the multi-agent literature within AI (e.g., see Bond & Gasser, 1988) when we assume that the agents involved are fully rational. As an aside, note that, in this case, we actually need not attribute much intelligence to the agents themselves, and it is sufficient to assume that their designers design them in a way that maximizes their own utility, disregarding the utility of the other agents.

In order to handle this problem we need to modify existing design paradigms. By adopting the perspective of partially controlled multi-agent systems, we obtain one possible handle on this problem, which requires making the following basic assumption: that the original designer, as in the above scenario, controls a number of reliable agents.[1] Our basic idea is that some of these reliable agents will be designed to *punish* agents that deviate from the desirable social standard. The punishment mechanism will be 'hard-wired' (unchangeable) and will be common-knowledge. The agents that are not controlled by the original designer will be aware of this punishment possibility. If the punishment mechanism is well designed, deviations from the social standard become irrational. As a result, *no* deviation will actually occur and *no* punishment will actually be executed! Hence, by making our agents a bit more sophisticated, we can prevent the temptation of breaking social laws.

In the suggested solution we adopt the perspective of partially controlled multi-agent systems. Some of the agents are controllable, while others are uncontrollable but are assumed to adopt the basic model of expected utility maximization. The punishment mechanism is (part of) the control strategy that is used to influence the behavior of the uncontrolled agents.

1. For ease of exposition, we assume that reliable agents follow the designer's instructions; we assume that no non-malicious failures, such as crash failures, are possible.

## 3. Dynamic Game Theoretic Model

In this section we introduce a basic game-theoretic model, which we use to study the problem of the enforcement of social behavior and its solution. Later on, in Sections 5–8, this model will be used to study embedded teaching. We wish to emphasize that the model we use is the most common model for representing emergent behavior in a population[2] (e.g., Huberman & Hogg, 1988; Kandori, Mailath, & Rob, 1991; Altenberg & Feldman, 1987; Gilboa & Matsui, 1991; Weidlich & Haag, 1983; Kinderman & Snell, 1980).

**Definition 1** *A $k$-*person game *$g$ is defined by a $k$-dimensional matrix $M$ of size $n_1 \times \cdots \times n_k$, where $n_m$ is the number of possible actions (or* strategies*) of the $m$'th agent. The entries of $M$ are vectors of length $k$ of real numbers, called* payoff vectors*. A* joint strategy *in $M$ is a tuple $(i_1, i_2, \ldots, i_k)$, where for each $1 \leq j \leq k$, it is the case that $1 \leq i_j \leq n_j$.*

Intuitively, each dimension of the matrix represents the possible actions of one of the $k$ players of the game. Following the convention used in game theory, we often use the term *strategy* in place of *action*. Since the dimensions of the matrix are $n_1 \times \cdots \times n_k$, the $i$'th agent has $n_i$ possible strategies to choose from. The $j$'th component of the vector residing in the $(i_1, i_2, \ldots, i_k)$ cell of $M$ (i.e., $M_{i_1,i_2,\ldots,i_k}$) represents the feedback player $j$ receives when the players' joint strategy is $(i_1, i_2, \ldots, i_k)$, that is, if agent $m$'s strategy is $i_m$ for all $1 \leq m \leq k$. Here, we use the term joint strategy to refer to the combined choice of strategies of all the agents.

**Definition 2** *A $n$-$k$-$g$ iterative game consists of a set of $n$ agents and a given $k$ person game $g$. The game $g$ is played repetitively an unbounded number of times. At each iteration, a random $k$-tuple of agents play an instance of the game, where the members of this $k$-tuple are selected with uniform distribution from the set of agents.*

Every iteration of an $n$-$k$-$g$ game represents some local interaction of $k$ agents. Those agents that play in a particular iteration of the game must choose the strategy they will use in this interaction; an agent can use different strategies in different interactions. The outcome of each iteration is represented by the payoff vector corresponding to the agents' joint strategy. Intuitively, this payoff tells us how good the outcome of this joint behavior is from the point of view of each agent. Many situations can be represented as an $n$-$k$-$g$ game, for example, the "traffic" aspect of a multi-agent system can be represented by an $n$-$k$-$g$ game, where each time a number of agents meet at an intersection. Each such encounter is an instance of a game in which agents can choose from a number of strategies, e.g., move ahead, yield. The payoff function gives the utility to each set of strategies. For example, if each time only two agents meet and both agents choose to move ahead, a collision occurs and their payoffs are very low.

**Definition 3** *A joint strategy of a game $g$ is called* efficient *if the sum of the players' payoffs is maximal.*

2. In this paper we use the term *emergent behavior* in its classical mathematical-economics interpretation: an evolution of a behavior based on repetitive local interactions of (usually pairs of) agents, where each agent may change its strategy for the following interactions based on the feedback it received in previous interactions.

Hence, efficiency is one global criterion for judging the "goodness" of outcomes from the system's perspective, unlike single payoffs which describe a single agent's perspective.[3]

**Definition 4** *Let $s$ be a fixed joint strategy for a given game $g$, with payoff $p_i(s)$ for player $i$; in an instance of $g$ in which a joint strategy $s'$ was played, if $p_i(s) \geq p_i(s')$ we say that $i$'s* punishment w.r.t. *$s$ is $p_i(s) - p_i(s')$, and otherwise we say that its* benefit w.r.t. *$s$ is $p_i(s') - p_i(s)$.*

Hence, punishment and benefit w.r.t. some joint strategy $s$ measure the gain (benefit) or loss (punishment) of an agent if we can somehow change the joint behavior of the agents from $s$ to $s'$.

In our current discussion punishment and benefit will always be with respect to a chosen efficient solution.

As designers of the multi-agent system, we would prefer it to be as efficient as possible. In some cases this entails behavior that is in some sense unstable, that is, individual agents may locally prefer to behave differently. Thus, agents may need to be constrained to behave in a way that is locally sub-optimal. We refer to such constraints that exclude some of the possible behaviors as *social laws.*

Due to the symmetry of the system and under the assumption that the agents are rational and their utility is additive (i.e., that the utility of two outcomes is the sum of their utilities), it is clear that no agent's expected payoff can be higher than the one obtained using the strategies giving the efficient solution. Thus, it is clear that in this case an efficient solution is fair, in the sense that all agents can get at least what they could if no such law existed, and no other solution can provide a better expected payoff.

However, the good intentions of the designer of creating an environment beneficial to the participating agents, may backfire. A social law provides information on the behavior of agents conforming to it, information that other agents (or their respective designers) can use to increase their expected payoff.

**Example 1** Assume that we are playing an $n$-2-$g$ game where g is the prisoner's dilemma, represented in strategic form by the following matrix.

| | agent 2 | |
|---|---|---|
| agent 1 | 1 | 2 |
| 1 | (2,2) | (-10,10) |
| 2 | (10,-10) | (-5,-5) |

The efficient solution of this game is obtained when both players play strategy 1. Assume that this solution is chosen by the original designer, and is followed by all agents under its control.

A designer of a new agent that will function in an environment in which the social law is obeyed may be tempted to program her agent not to conform to the chosen law. Instead, he will program the agent to play the strategy the maximizes its expected outcome, strategy

3. Addition of payoffs or utilities across agents is a dangerous practice. However, in our particular model, it can be shown that a system in which joint-strategies are always efficient maximizes each agent's expected cumulative rewards.

#2. This new agent will obtain a payoff of 10 when playing against one of the 'good' agents. Thus, even though the social law was accepted in order to guarantee a payoff of 2 to any agent, 'good' agents will obtain a payoff of -10 when playing against such non-conforming agents. Note that the new designer exploits information on the strategies of 'good' players, as dictated by the social law. The agents controlled by the new designer are uncontcolable agents; their behavior can not be dictated by the original designer. ∎

Agents not conforming to the social law will be referred to as *malicious agents*. In order to prevent the temptation to exploit the social law, we introduce a number of *punishing agents*, designed by the initial designer, that will play 'irrationally' if they detect behavior not conforming to the social law, attempting to minimize the payoff of the malicious agents. The knowledge that future participants have of the punishment policy would deter deviations and eliminate the need for carrying it out. Hence, the punishing behavior is used as a threat aimed at deterring other agents from violating the social law. This threat is (part of) the control strategy adopted the controllble agents in order to influence the behavior of the unconrollable agents. Notice that this control strategy relies on the structural assumption that the unconrollable agents are expected utility maximizers.

We define the *minimized malicious payoff* as the minimal expected payoff of the malicious players that can be guaranteed by the punishing agents. A punishment *exists*, if the minimized malicious payoff is lower than the expected payoff obtained by playing according to the social law. A strategy that guarantees the malicious agents an expected payoff lower than the one obtained by playing according to the social law is called a *punishing strategy*. Throughout this section and the following section we make the natural assumption that the expected payoff of malicious agents when playing against each other is no greater than the one obtained in the efficient solution[4].

**Example 1** (*continued*) In Example 1, the punishment would simply be to play strategy 2 from now on. This may cause the payoff of a punishing agent to decrease, but would guarantee that no malicious agent obtains a payoff better than -5 playing against a punishing agent. If many non-malicious agents are punishing, the malicious agents' expected payoff would decrease and become smaller then the payoff guaranteed by the social law. Strategy 2 would be the punishing strategy. ∎

## 4. The Design of Punishments

In the previous section we described a general model of multi-agent interaction and showed how the perspecive of partially controlled multi-agent systems leads to one possible solution to the problem of enforcing social behavior in this setting, via the idea of threats and punishments. We now proceed to examine the issue of punishment design.

We assume that there are $p$ agents which the designer controls that either have an ability to observe instances of the game that occur, or that can be informed as to the outcome of games. There are $c$ additional agents that conform with the law (that is, play the strategies entailed by the chosen efficient solution), and $m$ malicious agents, that are not bound by the law.

4. Other assumptions may be treated similarly.

We would like to answer questions such as: Does a game offer the ability to punish? What is the minimized malicious payoff? What is the optimal ratio between $p, c$, and $m$? Is there a difference between different social laws?

**Example 1** (*continued*) Consider Example 1 again. We have observed above that we can cause an expected maximal loss for the malicious agents of 7 $(= 2 - (-5))$. This occurs when the punishing agents play strategy 2. The gain that a malicious agent makes when playing against an agent following the social law is 8 $(= 10 - 2)$. In order for a punishing strategy to be effective, it must be the case that the expected payoff of a malicious agent will be no greater than the expected payoff obtained when following the social law. In order to achieve this, we must ensure that the ratio of punishing/conforming agents is such that a malicious agent will have sufficient encounters with punishing agents. In our case, assuming that when 2 deviators meet their expected benefit is 0 and recalling that an agent is equally likely to meet any other agent, we need $\frac{p}{c} > \frac{8}{7}$ to make the incentive to deviate negative. ∎

Implementing the punishment approach requires more complex behavior. Our agents must be able to detect deviations as well as to switch to a new punishing strategy. This whole behavior can be viewed as a new, more complex, social law. This calls for more complex agents to carry it out, and makes the programming task harder.

Clearly, we would like to minimize the number of such complex agents, keeping the benefit of malicious behavior negative. Here, the major question is the ratio between the benefit of deviation and the prospective punishment.

As can be seen from the example, the larger the punishment, the smaller the number of the more sophisticated punishing agents that is needed. Therefore, we would like to find out which strategies minimize the malicious agent's payoff. In order to do this we require a few additional definitions.

**Definition 5** *A two person game $g$ is a zero-sum game if for every joint strategy of the players, the sum of the players' payoffs is 0.*

Hence, in a zero-sum game, there are no win/win situations, the larger the payoff of one agent, the smaller the payoff of the other agent. By convention, the payoff matrix of a two person zero-sum game will mention only the payoffs of player 1.

**Definition 6** *Let $g$ be a two person game. Let $P_i^g(s,t)$ be the payoff of player $i$ in $g$ (where $i \in \{1,2\}$) when strategies $s$ and $t$ are played by player 1 and 2 respectively. The* projected game, *$g_p$, is the following two person zero-sum game: The strategies of both players are as in $g$, and the payoff matrix is $P^{g_p}(s,t) = -P_2^g(s,t)$. Define the* transposed *game of $g$, $g^T$, to be the game $g$ where the roles of the players change.*

In the projected game, the first agent's payoff equals the negated value of the second agent's payoff in the original game. Thus, this game reflects the desire to lower the payoffs of the second player in the original game.

We give a general result for a two-person game, $g$ (with any number of strategies). We make use of the following standard game-theoretic definition:

**Definition 7** *Given a game $g$, a joint strategy $\sigma$ for the players is a* Nash equilibrium *of $g$ if whenever a player takes an action that is different than its action at $\sigma$, its payoff given that the other players play as in $\sigma$ is no higher than its payoff given that everybody plays $\sigma$.*

That is, a strategy $\sigma$ is a Nash equilibrium of a game if no agent can obtain a better payoff by unilaterally changing its behavior when all the other agents play according to $\sigma$.

Nash-equilibrium is the central notion in the theory of non-cooperative games (Luce & Raiffa, 1957; Owen, 1982; Fudenberg & Tirole, 1991). As a result, this notion is well studied and understood, and reducing new concepts to this basic concept may be quite useful from a design perspective. In particular, Nash-equilibrium always exists for finite games, and the payoffs prescribed by any Nash-equilibria of a given zero-sum game are uniquely defined. We can show:

**Theorem 1** *Given an $n$-2-$g$ iterative game, the minimized malicious payoff is achieved by playing the strategy of player 1 prescribed by the Nash equilibrium of the projected game $g_p$, when playing player 1 (in $g$), and the strategy of player 1 prescribed by the Nash equilibrium of the projected game $(g^T)_p$, when playing player 2 (in $g$).*[5]

**Proof:** Assume that the punishing agent plays the role of player 1. If player 1 adopts the strategy prescribed by a Nash-equilibrium $\sigma$ then player 2 can not get a better payoff than the one guaranteed by $\sigma$ since each deviation by player 2 will not improve its situation (by the definition of Nash-equilibrium). On the other hand, player 1 can not cause more harm than the harm obtained by playing its strategy in $\sigma$. To see this, assume that player 1 uses an arbitrary strategy $s$, and that player 2 adopts the strategy prescribed by $\sigma$. The outcome for player 1 will be not higher than the one guaranteed by playing the Nash-equilibrium (by the definition of Nash-equilibrium). In addition, due to the fact that we have here a zero-sum game this implies that the outcome for player 2 will be no lower than the one guaranteed if player 1 would play according to $\sigma$. The case where the punishing agent is player 2 is treated similarly. ∎

**Example 1** (*continued*) Continuing our prisoner's dilemma example, $g_p$ would be

| agent 1 \ agent 2 | 1 | 2 |
|---|---|---|
| 1 | -2 | -10 |
| 2 | 10 | 5 |

with the Nash equilibrium attained by playing the strategies yielding 5. In this example, $(g^T)_p = g_p$. Therefore, the punishing strategies will be strategy # 2 for each case. ∎

**Corollary 1** *Let $n$-2-$g$ be an iterative game, with $p$ punishing agents. Let $v$ and $v'$ be the payoffs of the Nash equilibria of $g_p$ and $g_p^T$ respectively (which, in this case, are uniquely defined). Let $b,b'$ be the maximal payoffs player 1 can obtain in $g$ and $g^T$ respectively,*

5. Notice that, in both cases, the strategies prescribed for the original game are determined by the strategies of player 1 in the Nash-Equilibria of the projected games.

*assuming player 2 is obeying the social law. Let e and e' be the payoffs of player 1 and 2, respectively, in g, when the players play according to the efficient solution prescribed by the social law. Finally, assume that expected benefit of two malicious agents when they meet is 0. A necessary and sufficient condition for the existence of a punishing strategy is that* $\frac{(n-1-p)}{n-1} \cdot (b+b') - \frac{p}{n-1} \cdot (v+v') < (e+e')$.

**Proof:** The expected payoff obtained by a malicious agent when encountering a law-abiding agent is $\frac{b+b'}{2}$, and its expected payoff when encountering a punishing agent is $\frac{-(v+v')}{2}$. In order to test the conditions for the existence of a punishing strategy we would need to consider the best case scenario from the point of view of a malicious agent; in such a case all non-punishing agents are law-abiding agents. In order to obtain the expected utility for a malicious agent we have to make an average of the above quantities taking into account the proportion of law-abiding and punishing agents in the population. This gives us that the expected utility for a malicious agent is $\frac{(n-1-p)}{2(n-1)} \cdot (b+b') - \frac{p}{2(n-1)} \cdot (v+v')$. By definition, a punishing strategy exists if and only if this expected utility is lower than the expected utility guaranteed by the social law. Since the expected utility which can be guaranteed by a social law is $\frac{e+e'}{2}$, we get the desired result. ∎

The value of the punishment, $\frac{(v+v')}{2}$ in the above, is independent of the efficient solution chosen, and $e+e'$ is identical for all efficient solutions, by definition. However, $b+b'$ depends on our choice of an efficient solution. When a number of such solutions exist, minimizing $b+b'$ is an important consideration in the design of the social law, as it affects the incentive to 'cheat'.

**Example 2** Let's look at a slightly different version of the prisoner's dilemma. The game matrix is

| | agent 2 | |
|---|---|---|
| agent 1 | 1 | 2 |
| 1 | (0,0) | (-10,10) |
| 2 | (10,-10) | (-5,-5) |

Here there are 3 efficient solutions, given by the joint strategies (1,1), (1,2), (2,1). In the case of (1,1) we have b+b'=20 (gained by playing strategy 2 instead of 1). In the case of (2,1) and (1,2) b+b'=5.

Clearly, there is more incentive to deviate from a social law prescribing strategies (1,1) than from a social law prescribing (2,1) or (1,2). ∎

To summarize, the preceding discussion suggests designing a number of punishing agents, whose behavior in punishment mode is prescribed by Theorem 1 in the case of $n$-2-$g$ games. By ensuring a sufficient number of such agents we take away any incentive to deviate from the social laws. Hence, given that the malicious agents are rational, they will follow the social norm, and consequently, there will be no need to utilize the punishment mechanism. We observed that different social laws leading to solutions that are equally efficient have different properties when it comes to punishment design. Consequently, under the assumption that we would like to minimize the number of punishing agents while guaranteeing an efficient solution to the participants, we should choose an efficient solution that minimizes the value of $b + b'$.

## 5. Embedded Teaching

In this section we move on to our second study of a PCMAS design problem; only now, the uncontrollable agent is a reinforcement learner. This choice is not arbitrary; rational agents and reinforcement learners are the two major types of agents studied in mathematical economics, decision theory, and game theory. They are also the types of agents discussed in work in DAI which is concerned with self-motivated agents (e.g., Zlotkin & Rosenschein, 1993; Kraus & Wilkenfeld, 1991; Yanco & Stein, 1993; Sen, Sekaran, & Hale, 1994).

An agent's ability to function in an environment is greatly affected by its knowledge of the environment. In some special cases, we can design agents with sufficient knowledge for performing a task (Gold, 1978), but, in general, agents must acquire information on-line in order to optimize their performance, i.e., they must learn. One possible approach to improving the performance of learning algorithms is employing a teacher. For example, Lin (1992) uses teaching by example to improve the performance of agents, supplying them with examples that show how the task can be achieved. Tan's work (1993) can also be viewed as a form of teaching in which agents share experiences. In both methods some non-trivial form of communication or perception is required. We strive to model a broad notion of teaching that encompasses any behavior that can improve a learning agent's performance. That is, we wish to conduct a general study of partially controlled multi-agent systems in which the uncontrollable agent runs a learning algorithm. At the same time, we want our model to clearly delineate the limits of the teacher's (i.e., the controlling agent's) ability to influence the student.

Here, we propose a teaching approach that maintains a situated "spirit" much like that of reinforcement learning (Sutton, 1988; Watkins, 1989; Kaelbling, 1990), which we call *embedded teaching*. An embedded teacher is simply a "knowledgeable" controlled agent situated with the student in a shared environment. Her[6] goal is to lead the student to adopt some specific behavior. However, the teacher's ability to teach is restricted by the nature of the environment they share: not only is her repertoire of actions limited, but she may also lack full control over the outcome of these actions. As an example, consider two mobile robots without any means of direct communication. Robot 1 is familiar with the surroundings, while Robot 2 is not. In this situation, Robot 1 can help Robot 2 reach its goal through certain actions, such as blocking Robot 2 when it is headed in the wrong direction. However, Robot 1 may have only limited control over the outcome of such an interaction because of uncertainty about the behavior of Robot 2 and its control uncertainty. Nevertheless, Robot 2 has a specific structure, it is a learner obeying some learning scheme, and we can attempt to control it indirectly through our choice of actions for Robot 1.[7]

6. To differentiate between teacher and student, we use female pronouns for the former and male pronouns for the latter.
7. In general, the fact that an agent is controllable does not imply that we can perfectly control the outcome of its actions, only their choice. Hence, a robot may be controllable in our sense, running a program supplied by us, yet its move-forward command may not always have the desired outcome.

In what follows, our goal is to understand how an embedded teacher can help a student adopt a particular behavior. We address a number of theoretical questions relating to this problem, and then we experimentally explore techniques for teaching two types of reinforcement learners.

## 6. A Basic Teaching Setting

We consider a teacher and a student that repeatedly engage in some joint activity. While the student has no prior knowledge pertaining to this activity, the teacher understands its dynamics. In our model, the teacher's goal is to lead the student to adopt a particular behavior in such interactions. For example, teacher and student meet occasionally at the road and the teacher wants to teach the student to drive on the right side. Or perhaps, the teacher and the student share some resource, such as CPU time, and the goal is to teach him judicious use of this resource. We model such encounters as 2-2-$g$ iterative games.

To capture the idea that the teacher is more knowledgeable than the student, we assume that she knows the structure of the game, i.e., she knows the payoff function, and that she recognizes the actions taken at each play. On the other hand, the student does not know the payoff function, although he can perceive the payoff he receives. In this paper, we make the simplifying assumptions that both teacher and student have only two actions from which to choose and that the outcome depends only on their choice of actions. Furthermore, excluding our study in Section 8.4, we ignore the cost of teaching, and hence, we omit the teacher payoff from our description.[8] This provides a basic setting in which to take this first step towards understanding the teaching problem.[9]

The teaching model can be concisely modeled by a $2 \times 2$ matrix. The teacher's actions are designated by $I$ and $II$, while the student's actions are designated by the numbers 1 and 2. Each entry corresponds to a joint action and represents the student's payoff when this joint action is played. We will suppose that we have matrix A of Figure 1, and that we wish to teach the student to use action 1. At this stage, all we assume about the student is that if he *always* receives a better payoff following action 1 he will learn to play it.

We can see that in some situations teaching is trivial. Assume that the first row dominates the second row, i.e., $a > c$ and $b > d$. In that case, the student will naturally prefer to take action 1, and teaching is not very challenging, although it might be useful in speeding the learning process. For example, if $a - c > b - d$, as in matrix B in Figure 1, the teacher can make the advantage of action 1 more noticeable to the student by always playing action $I$.

Now suppose that only one of $a > c$ or $b > d$ holds. In this case, teaching is still easy. We use a basic teaching strategy, which we call *preemption*. In preemption the teacher chooses an action that makes action 1 look better than action 2. For example, when the situation is described by matrix C in Figure 1, the teacher will always choose action $I$.

8. A case could be made for the inherent value of teaching, but this may not be the appropriate forum for airing these views.
9. In fact, our idea has been to consider the most basic embedded teaching setting which is already challenging. As we later see, this basic setting is closely related to a fundamental issue in non-cooperative games.

| | *I* | *II* |
|---|---|---|
| **1** | a | b |
| 2 | c | d |

(A)

| | *I* | *II* |
|---|---|---|
| **1** | 6 | 5 |
| 2 | 1 | 2 |

(B)

| | *I* | *II* |
|---|---|---|
| **1** | 5 | 1 |
| 2 | 2 | 6 |

(C)

| | *I* | *II* |
|---|---|---|
| **1** | 3 | -2 |
| 2 | 5 | 6 |

(D)

| | *I* | *II* |
|---|---|---|
| **1** | 5 | -10 |
| 2 | 10 | -5 |

(E)

Figure 1: Game matrices A, B, C, D, and E. The teacher's possible actions are $I$ and $II$, and the student's possible actions are 1 and 2.

Next, assume that both $c$ and $d$ are greater than both $a$ or $b$, as in matrix D in Figure 1. Regardless of which action the teacher chooses, the student receives a higher payoff by playing action 2 (since $\min\{5, 6\} > \max\{3, -2\}$). Therefore, no matter what the teacher does, the student will learn to prefer action 2. Teaching is hopeless in this situation.

All other types of interactions are isomorphic to the case where $c > a > d > b$, as in matrix E in Figure 1. This is still a challenging situation for the teacher because action 2 dominates action 1 (because $10 > 5$ and $-5 > -10$). Therefore, preemption cannot work. If a teaching strategy exists, it will be more complex than always choosing the same action. Since this seems to the most challenging teaching situation, we devote our attention to teaching a reinforcement learner to choose action 1 in this class of games.

It turns out that the above situation is quite important in game-theory and multi-agent interaction. It is a projection of a very famous game, the prisoner's dilemma, discussed in the previous sections. In general, we can represent the prisoner's dilemma using the following game matrix:

| student \ teacher | Coop | Defect |
|---|---|---|
| Coop | (a,a) | (b,c) |
| Defect | (c,b) | (d,d) |

or more commonly

| student \ teacher | Coop | Defect |
|---|---|---|
| Coop | (a,a) | (-c,c) |
| Defect | (c,-c) | (d,d) |

where $c > a > d > b$. The actions in the prisoner's dilemma are called Cooperate (Coop) and Defect; we identify Coop with actions 1 and $I$, and Defect with actions 2 and $II$. The prisoner's dilemma captures the essence of many important social and economic situations; in particular, it encapsulates the notion of cooperation. It has thus motivated enormous discussion among game-theorists and mathematical economists (for an overview, see Eatwell, Milgate, & Newman, 1989). In the prisoner's dilemma, whatever the choice of one player, the second player can maximize its payoff by playing Defect. It thus seems "rational" for each player to defect. However, when both players defect, their payoffs are much worse than if they both cooperate.

As an example, suppose two agents will be given \$10 each for moving some object. Each agent can perform the task alone, but it will take an amount of time and energy which they value at \$20. However, together, the effort each will make is valued at \$5. We get the following instance of the prisoner's dilemma:

| | Agent 1 | |
|---|---|---|
| Agent 2 | Move | Rest |
| Move | (5,5) | (-10,10) |
| Rest | (10,-10) | (0,0) |

In the experimental part of our study, the teacher's task will be to teach the student to cooperate in the prisoner's dilemma. We measure the success of a teaching strategy by looking at the cooperation rate it induces in students over some period of time, that is, the percentage of the student's actions which are Coop. The experimental results presented in this paper involving the prisoner's dilemma are with respect to the following matrix:

| | Teacher | |
|---|---|---|
| Student | Coop | Defect |
| Coop | (10,10) | (-13,13) |
| Defect | (13,-13) | (-6,-6) |

We have observed qualitatively similar results in other instantiations of the prisoner's dilemma, although the precise cooperation rate varies.

## 7. Optimal Teaching Policies

In the previous section we concentrated on modeling the teaching context as an instance of a partially controlled multi-agent system, and determining which particular problems are interesting. In this section we start exploring the question of how a teacher should teach. First, we define what an optimal policy is. Then, we will define Markov decision processes (MDP) (Bellman, 1962), and show that under certain assumptions teaching can be viewed as an MDP. This will allow us to tap into the vast knowledge that has accumulated on solving these problems. In particular, we can use well known methods, such as value iteration (Bellman, 1962), to find the optimal teaching policy.

We start by defining an optimal teaching policy. A *teaching policy* is a function that returns an action at each iteration; possibly, it may depend on a complete history of the past joint actions. There is no "right" definition for an optimal policy, as the teacher's motivation may vary. However, in this paper, the teacher's objective is to maximize the number of iterations in which the student's action are "good", such as Coop in the prisoner's dilemma. The teacher does not know the precise number of iterations she will be playing, so she slightly prefers earlier success to later success.

This is formalized as follows: Let $u(a)$ be the value the teacher places on a student's action, $a$, let $\pi$ be the teacher's policy, and assume that it induces a probability distribution $Pr_{\pi,k}$ over the set of possible student actions at time $k$. We define the value of the strategy $\pi$ as

$$val(\pi) = \sum_{k=0}^{\infty} \gamma^k E_k(u)$$

where $E_k(u)$ is the expected value of $u$:

$$E_k(u) = \sum_{a \in A_s} Pr_{\pi,k}(a) \cdot u(a)$$

Here, $A_s$ is the student's set of actions. The teacher's goal is to find a strategy $\pi$ that maximizes $val(\cdot)$, the discounted expected value of the student's actions. For example, in the case of the prisoner's dilemma, we could have
$A_s = \{\text{Coop,Defect}\}$ and $u(\text{Coop}) = 1$ and $u(\text{Defect}) = 0$.

Next, we define MDPs. In an MDP, a decision maker is continually moving between different states. At each point in time she observes her current state, receives some payoff (which depends on this state), and chooses an action. Her action and her current state determine (perhaps stochastically) her next state. The goal is to maximize some function of the payoffs. Formally, an MDP is a four-tuple $\langle S, A, P, r\rangle$, where $S$ is the state-space, $A$ is the decision-maker's set of possible actions, $P : S \times S \times A \rightarrow [0, 1]$ is the probability of a transition between states given the decision-maker's action, and $r : S \rightarrow \Re$ is the reward function. Notice that given an initial state $s \in S$, and a policy of the decision maker $\pi$, $P$ induces a probability distribution $P_{s,\pi,k}$ over $S$, where $P_{s,\pi,k}(s')$ is the probability that the $k^{th}$ state obtained will be $s'$ if the current state is $s$.

The $\gamma_0$-optimal policy in an MDP is the policy that maximizes at each state $s$ the discounted sum of the expected values of the payoffs received at all future states, starting at $s$, i.e.,

$$\sum_{k=0}^{\infty} \gamma_0^k \left(\sum_{s' \in S} P_{s,\pi,k}(s') \cdot r(s')\right)$$

Although it may not be immediately obvious, a single policy maximizing discounted sums for any starting state exists, and there are well-known ways of finding this policy. In the experiments below we use a method based on value-iteration (Bellman, 1962).

Now suppose that the student can be in a set $\Sigma$ of possible states, that his set of actions is $A_s$, and that the teacher's set of actions is $A_t$. Moreover, suppose that the following properties are satisfied:
(1) The student's new state is a function of his old state and the current joint-action, denoted by $\tau : \Sigma \times A_s \times A_t \rightarrow \Sigma$;
(2) The student's action is a stochastic function of his current state, where the probability of choosing $a$ at state $s$ is $\rho(s, a)$;
(3) the teacher knows the student's state. (The most natural way for this to happen is that the teacher knows the student's initial state, the function $\tau$, and the outcome of each game, and she uses them to simulate the agent.)

Notice that under these assumptions a teaching policy should be a function of $\Sigma$: We know that the student's next action is a function of his next state. We know that the student's next state is a function of his current state, his current action, and the teacher's current action. Hence, his next action is a function of his current state and action, as well as the teacher's current action. However, we know that the student's current action is a function of his current state. Hence, the student's next action is a function of his current state and the teacher's current action. This implies that the only knowledge the teacher needs to optimally choose her current action is the student's current state, and any additional information will be redundant and cannot improve her success. More generally, when we repeat this line of reasoning indefinitely into the future, we see that the teacher's policy should be a function of the student's state: a function from $\Sigma$ to $A_t$. It is now possible to see that we have the makings of the following MDP.

Given this observation and our three assumptions, we see that, indeed, the teacher's policy induces a probability distribution over the set of possible student actions at time $k$. This implies that our definition of *val* makes sense here.

Define the teacher's MDP to be TMDP= $\langle \Sigma, A_t, P, U \rangle$, where

$$P(s, s', a_t) \stackrel{\text{def}}{=} \sum_{a_s \in A_s} \rho(s, a_s) \cdot \delta_{s', \tau(s, a_s, a_t)}$$

($\delta_{i,j}$ is defined as 1 when $i = j$, and 0 otherwise). That is, the probability of a transition from $s$ to $s'$ under $a_t$ is the sum of probabilities of the student's actions that will induce this transition. The reward function is the expected value of $u$:

$$U(s) \stackrel{\text{def}}{=} \sum_{a_s \in A_s} \rho(s, a_s) \cdot u(a_s)$$

**Theorem 2** *The optimal teaching policy is given by the $\gamma_0$ optimal policy in TMDP.*

**Proof:** By definition, the $\gamma_0$ optimal policy in TMDP is the policy $\pi$ that for each $s \in \Sigma$ maximizes

$$\sum_{k=0}^{\infty} \gamma_0^k \left( \sum_{s' \in \Sigma} P_{s,\pi,k}(s') \cdot U(s') \right)$$

that is,

$$\sum_{k=0}^{\infty} \gamma_0^k \left( \sum_{s' \in \Sigma} P_{s,\pi,k}(s') \cdot \left( \sum_{a_s \in A_s} \rho(s', a_s) \cdot u(a_s) \right) \right)$$

However, this is equal to

$$(*) \quad \sum_{k=0}^{\infty} \gamma_0^k \sum_{a_s \in A_s} \sum_{s' \in \Sigma} \rho(s', a_s) \cdot P_{s,\pi,k}(s') \cdot u(a_s)$$

We know that $P_{s,\pi,k}(s')$ is the probability that $s'$ will be the state of the student in time $k$, given that the teacher uses $\pi$ and that her current state is $s$. Hence,

$$\sum_{s' \in \Sigma} \rho(s', a_s) \cdot P_{s,\pi,k}(s')$$

is the probability that $a_s$ will be the action taken by the student at time $k$ given the initial (current) state is $s$. Upon examination, we see now that (*) is identical to $val(\pi)$. ∎

The optimal policy can be used for teaching, when the teacher possess sufficient information to determine the current state of the student. But even when that is not the case, it allows us to calculate an upper bound on the success $val(\pi)$ of any teaching policy $\pi$. This number is a property of the *learning* algorithm, and measures the degree of influence any agent can have over the given student.

## 8. An Experimental Study

In this section we describe an experimental study of embedded teaching. First, we define the learning schemes considered, and then, we describe a set of results obtained using computer simulations.

### 8.1 The Learning Schemes

We experiment with two types of students: One uses a reinforcement learning algorithm which can be viewed as Q-learning with one state, and the other uses Q-learning. In choosing parameters for these students we tried to emulate choices made in the reinforcement learning literature.

The first student, which we call a *Blind Q-learner* (BQL), can perceive rewards, but cannot see how the teacher has acted or remember his own past actions. He only keeps one value for each action, for example, $q(\text{Coop})$ and $q(\text{Defect})$ in the case of the prisoner's dilemma. His update rule is the following: if he performed action $a$ and received a reward of $R$ then

$$q_{new}(a) = (1-\alpha)\cdot q_{old}(a) + \alpha \cdot R$$

The parameter $\alpha$, the *learning-rate*, is fixed (unless stated otherwise) to 0.1 in our experiments. We wish to emphasize that although BQL is a bit less sophisticated than "real" reinforcement learners discussed in the AI literature (which is defined below), it is a popular and powerful type of learning rule, which is much discussed and used in the literature (Narendra & Thathachar, 1989).

The second student is a Q-learner (QL). He can observe the teacher's actions and has a number of possible states. The QL maintains a Q-value for each state-action pair. His states encode his recent experiences, i.e., the past joint actions. The update rule is:

$$q_{new}(s,a) = (1-\alpha)\cdot q_{old}(s,a) + \alpha \cdot (R + \gamma V(s'))$$

Here $R$ is the reward received upon performing $a$ at state $s$; $s'$ is the state of the student following the performance of $a$ at $s$; $\gamma$ is called the *discount factor*, and will be 0.9, unless otherwise noted; and $V(s')$ is the current estimate of the value of the best policy on $s'$, defined as $\max_{a\in A_s} q(s',a)$. All Q-values are initially set to zero.

The student's update rule tells us how his Q-values change as a result of new experiences. We must also specify how these Q-values determine his behavior. Both $QL$ and $BQL$ students choose their actions based on the Boltzmann distribution. This distribution associates a probability $P_s(a)$ with the performance of an action $a$ at a state $s$ ($P(a)$ for the BQL).

$$P_s(a) \stackrel{\text{def}}{=} \frac{\exp(q(s,a)/T)}{\sum_{a'\in A}\exp(q(s,a')/T)} \quad (QL) \qquad P(a) \stackrel{\text{def}}{=} \frac{\exp(q(a)/T)}{\sum_{a'\in A}\exp(q(a')/T)} \quad (BQL)$$

Here $T$ is called the *temperature*. Usually, one starts with a high value for $T$, which makes the action choice more random, inducing more exploration on the part of the student. $T$ is slowly reduced, making the Q-values play a greater role in the student's choice of action. We use the following schedule: $T(0) = 75$ and $T(n+1) = T(n)*0.9+0.05$. This schedule has the characteristic properties of fast initial decay and slow later decay. We also experiment with fixed temperature.

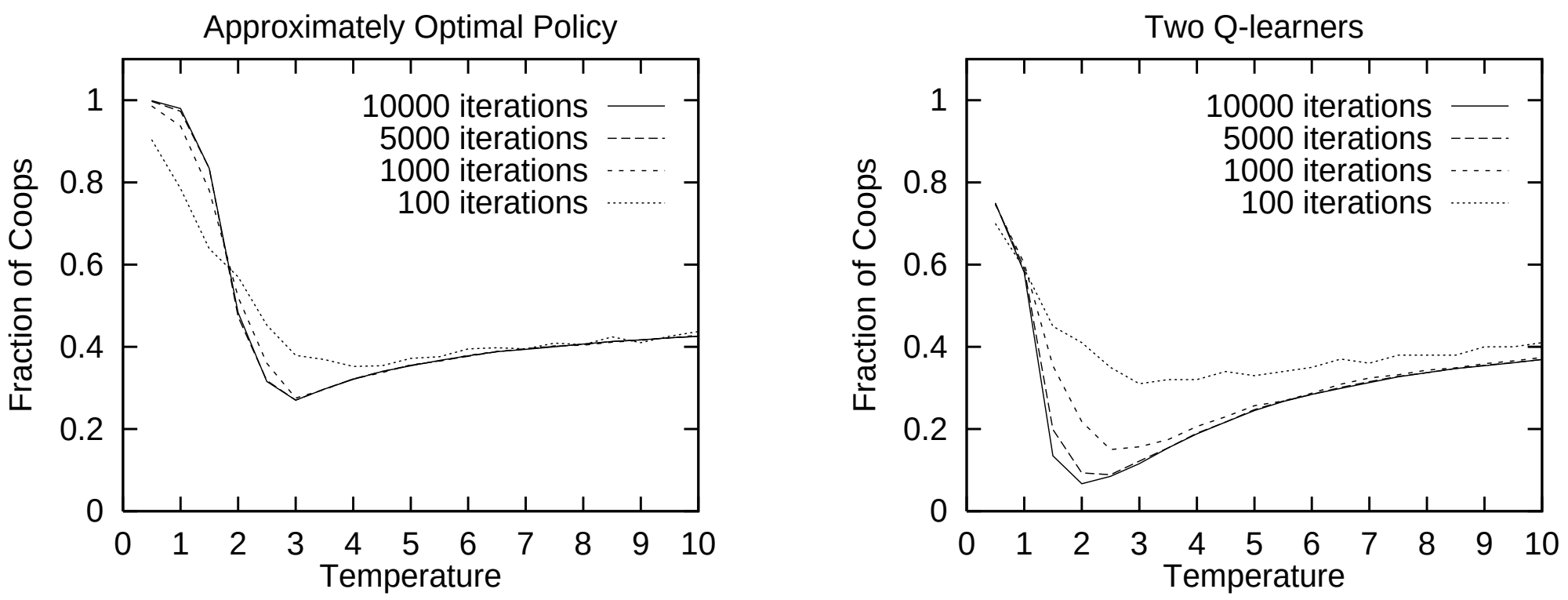

Figure 2: Fraction of Coops as a function of temperature for the approximately optimal policy (left) and for "teaching" using an identical Q-learner (right). Each curve corresponds to Coop rate over some fixed number of iterations. In the approx. optimal policy the curves for 1000, 5000 and 10000 iterations are nearly identical.

### 8.2 Blind Q-Learners

Motivated by our discussion in Section 6 we will concentrate in this section and in the following section on teaching in the context of the prisoner's dilemma. In Section 8.4 we discuss another type of teaching setting. This section describes our experimental results with BQL. We examined a policy that approximates the optimal policy, and two teaching methods that do not rely on a student model.

#### 8.2.1 Optimal Policy

First we show that BQLs fit the student model of Section 7. For their state space, we use the set of all possible assignment for their Q-values. This is a continuous subspace of $\Re^2$, and we discretize it (in order to be able to compute a policy), obtaining a state space with approximately 40,000 states. Next, notice that transitions are a stochastic function of the current state (current Q-values) and the teacher's action. To see this notice that Q-value updates are a function of the current Q-value and the payoff; the payoff is a function of the teacher's and student's actions; and the student's actions are a stochastic function of the current Q-value. In the left side of Figure 2 we see the success of teaching using the policy generated by using dynamic programming to solve this optimization problem. Each curve represents the fraction of Coops as a function of the temperature for some fixed number of iterations. The values are means over 100 experiments.

#### 8.2.2 Two Q-Learners

We also ran experiments with two identical BQLs. This can be viewed as "teaching" using another Q-learner. The results are shown in the right side of Figure 2. At all temperatures the optimal strategy performs better than Q-learning as a "teaching" strategy. The fact that at temperatures of 1.0 or less the success rate approaches 1 will be beneficial when we later add temperature decay. However, we also see that there is an inherent limit to our ability to affect the behavior at higher temperatures. An interesting phenomenon is the phase transition observed around $T = 2.5$. A qualitative explanation of this phenomenon is that high temperature adds randomness to the student's choice of action, because it makes the probabilities $P(a)$ less extreme. Consequently, the ability to predict the student's behavior lessens, and with it the probability of choosing a good action. However, while randomness serves to lower the success rate initially, it also guarantees a level of effective cooperation, which should approach 0.5 as the temperature increases. Finally, notice that although (Coop,Coop) seems like the best joint-action for a pair of agents, two interacting Q-learners never learn to play this joint strategy consistently, although they approach 80% Coops at low temperatures.

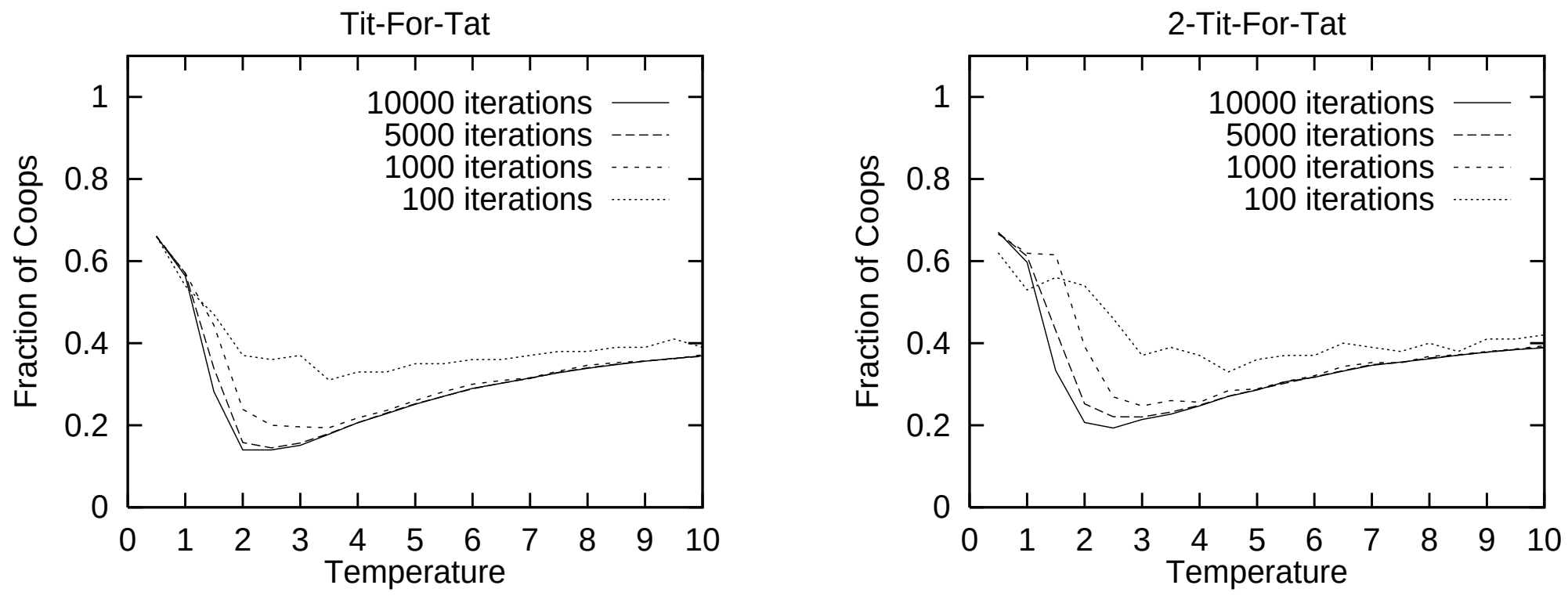

Figure 3: Fraction of Coops as a function of temperature for the teaching strategy based on TFT (left) and 2TFT (right).

#### 8.2.3 Teaching Without a Model

When the teacher does not have a precise model of the student, we cannot use the techniques of Section 7 to derive an optimal policy; in these models, we assume that the teacher can "observe" the student's current state (i.e. that it knows the student's Q-values). We therefore explore two teaching methods that only exploit knowledge of the game and the fact that the student is a BQL.

Both methods are motivated by a basic strategy of *countering* the student's move. The basic idea is to try and counter good actions by the student with an action that will lead to a high payoff, and to counter bad actions with an action that will give him a low payoff. Ideally, we would like to play Coop when the student plays Coop, and Defect when the student plays Defect. Of course, we don't know what action the student will choose, so we try to predict from his past actions.

If we assume that the Q-values change very little from one iteration to the other, the student's most likely action in the next game is the same action that he took in the most recent game. Therefore, if we saw the student play Coop in the *previous* turn, we will play Coop *now*. Similarly, the teacher will follow a Defect by the student with a Defect on her part. This strategy, called Tit-For-Tat (TFT for short), is well known (Eatwell et al., 1989). Our experiments show that it is not very successful in teaching a BQL (see Figure 3).

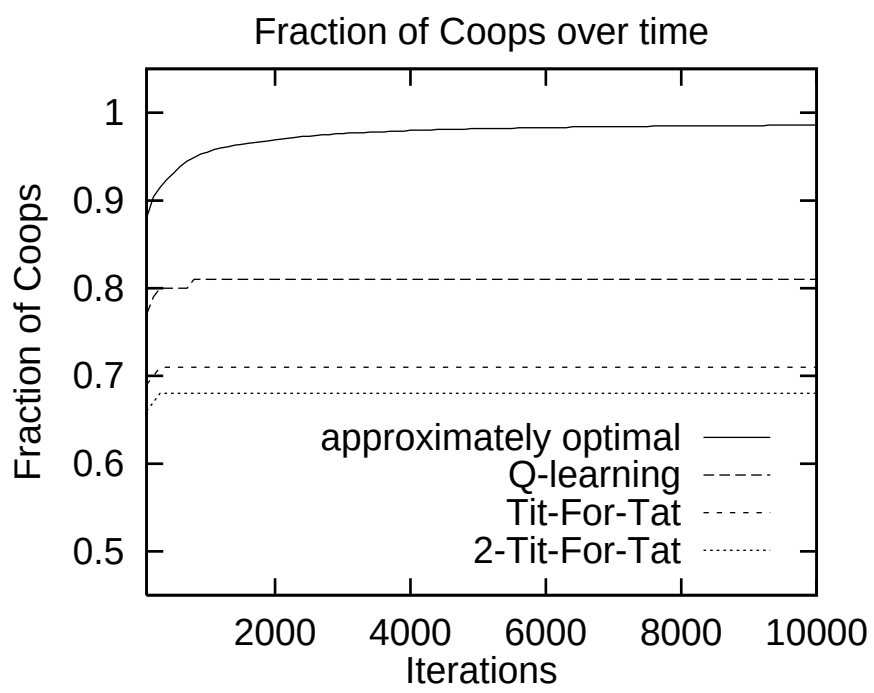

Figure 4: Fraction of Coops as a function of time for BQL using the temperature decay scheme of Section 8.1. Teaching strategies shown: approximately optimal strategy, Q-learning, TFT, and 2TFT.

We also experimented with a variant of TFT, which we call 2TFT. In this strategy the teacher plays Defect only after observing two consecutive Defects on the part of the student. It is motivated by our observation that in certain situations it is better to let the student enjoy a free lunch (that is, match his Defect with a Coop) than to make Coop look bad to him, because that may cause his Q-value for Coop to be so low that he is unlikely to try it again. Two consecutive Defects indicate that the probability of the student playing Defect next is quite high. The results, shown in Figure 3, indicate that this strategy worked better than TFT, and in some ranges of temperature, better than Q-learning. However, in general, both TFT and 2TFT gave disappointing results.[10]

Finally, Figure 4 shows the performance of all four teaching strategies discussed so far when we incorporate temperature decay. We can see that the optimal policy is very successful. As we explained before, teaching is easier when the student is more predictable, which is the case when temperature is lower. With temperature decay the student spends most of his time in relatively low temperature and behaves similarly to the case of fixed, low temperature. While an initial high-temperature phase could have altered this behavior, we did not observe such effects.

### 8.3 Teaching Q-Learners

Unlike BQL, Q-learners (QL) have a number of possible states which encode the joint actions of previous games played. A QL with memory one has four possible states, corresponding to the four possible joint actions in the prisoner's dilemma; a QL with more memory will have more states, encoding a sequence of joint actions.

More complex learning architectures have more structure, which brings with it certain problems. One possible problem may be that this structure is more "teaching-resistant." A

10. In some sense the use of an identical Q-learner implies having a model of the student, while TFT and 2TFT do not make use of such a model.

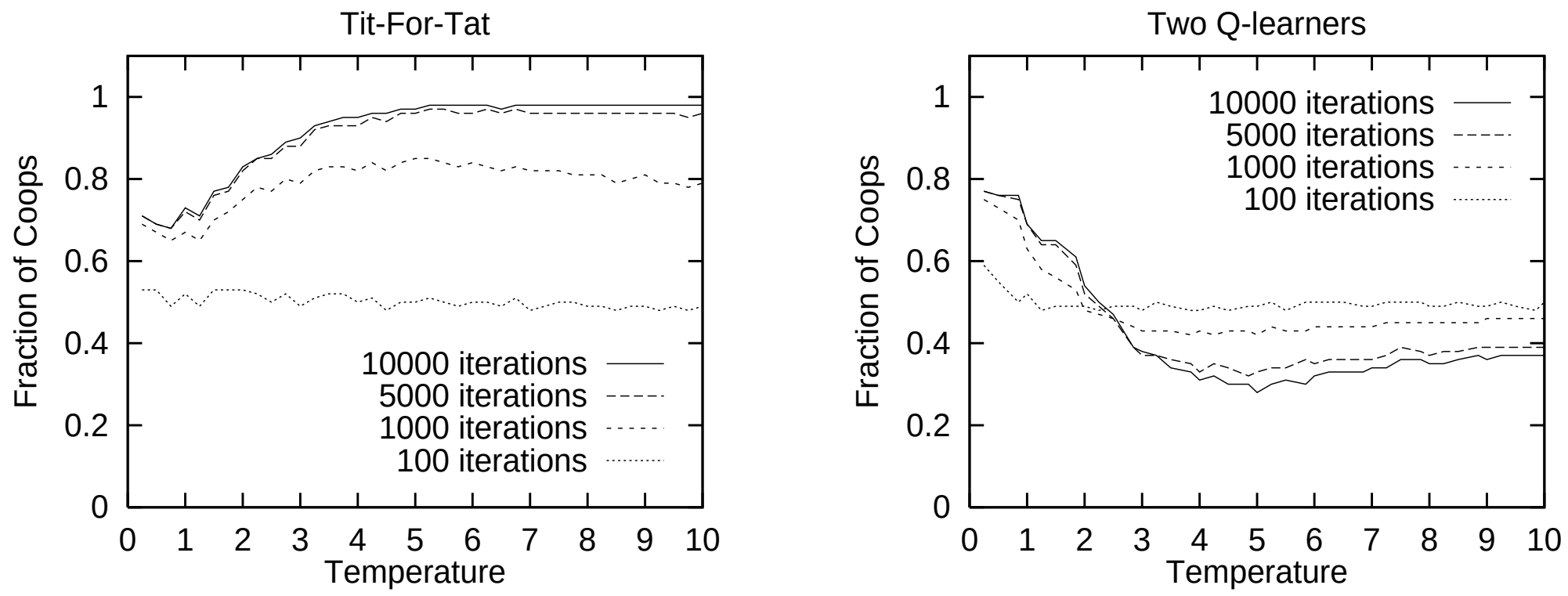

Figure 5: Each curve shows the fraction of Coops of QL as a function of temperature for a fixed number of iterations when TFT was used to teach (left) and when an identical Q-learner was used to teach (right). Values are means over 100 experiments.

more real threat is added computational complexity. As we mentioned, to approximate the optimal teaching policy for BQL we had to compute over a space of approximately 40,000 discretized states. While representing the state of a BQL requires only two numbers, one for each Q-value, representing the state of a QL with $m$ states requires $2m + 1$ numbers: one for the Q-value of each state/action pair, and one encoding the current state. The size of the corresponding discretized state-space for the teacher's Markov decision process grows exponentially in $m$. For the simplest case of memory one (a student with four states) this would be about $10^{18}$ states. Since solving the problem with 40,000 states took 12 hours on a SUN SPARCSTATION-10, we were not able to approximate optimal teaching policies for even the simplest QL.

But all is not lost. More structure may mean more complexity, but it also means more properties to exploit. We can reach surprisingly good results by exploiting the structure of Q-learners. Moreover, we can do this using a teaching method introduced in the previous section. However, in QL this method takes on a new meaning that suggests the familiar notions of reward and punishment. Interestingly, one may recall that punishment has been our major tool in our approach to the enforcement of social behavior.

In choosing their actions, QLs "care" not only about immediate rewards, but also about the current action's effect on future rewards. This makes them suitable for a reward and punishment scheme. The idea is the following: suppose the QL did something "bad" (Defect in our case). Although we cannot reliably counter such a move with a move that will lower his reward, we can punish him later by choosing an action that always gives a negative payoff, no matter what the student plays. We achieve this by following a student's Defect with a Defect by the teacher. While the immediate reward obtained by a QL playing Defect may be high, he will also learn to associate a subsequent punishment with the Defect action. Thus, while it may be locally beneficial to perform Defect, we may be able to make the long-term rewards of Defect less desirable. Similarly, we can follow a student's Coop with a reward in the form of a Coop by the teacher, since it guarantees a positive payoff to the student.

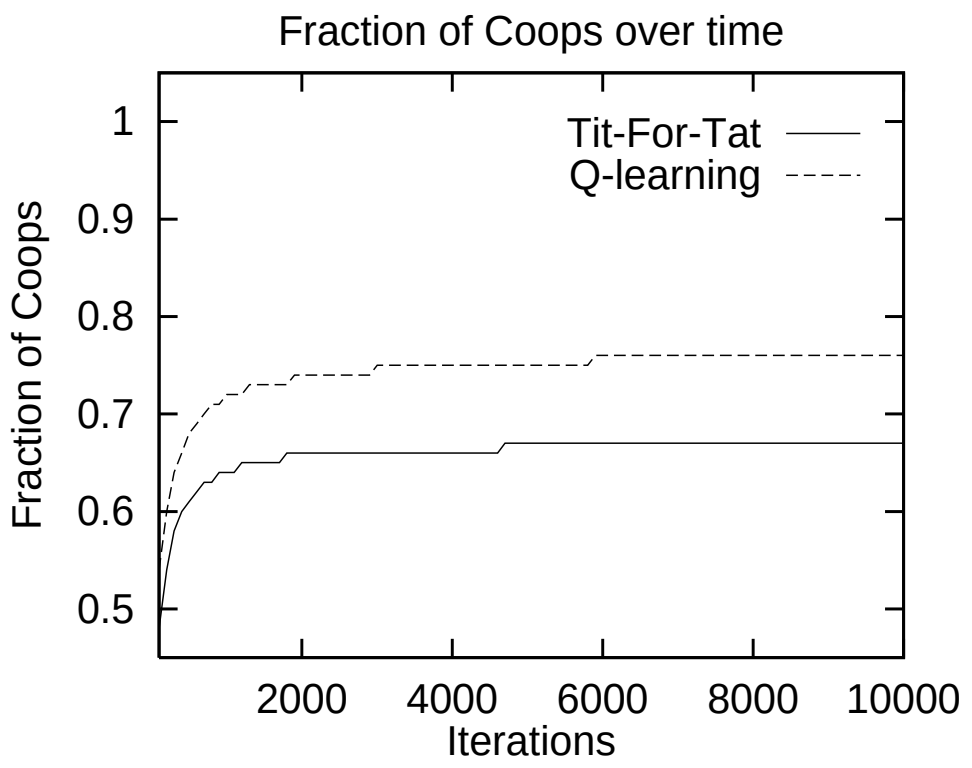

Figure 6: Fraction of Coops of QL as a function of time with temperature decay with TFT and with Q-learning as teaching strategies.

This suggests using Tit-For-Tat again. Notice that for BQLs, TFT *cannot* be understood as a reward/punishment strategy because BQLs care only about the immediate outcome of an action; the value they associate with each action is a weighted average of the immediate payoffs generated by playing this action.

In Figure 5 we see the success rates of TFT as a function of temperature, as well as the rates for Q-learning as a teaching strategy. In this latter case, the teacher is identical to the student. It is apparent that TFT is extremely successful, especially in higher temperatures. Interestingly, the behavior is quite different than that of two QLs. Indeed, when we examine the behavior of two QLs, we see that, to a lesser extent, the phase change noticed in BQLs still exists. We obtain completely different behavior when TFT is used: Coop levels increase with temperature, reaching almost 100% above 3.0. Hence, we see that TFT works better when the student Q-learner exhibits a certain level of experimentation. Indeed, if we examine the success of these teaching strategies at a very low temperature, we see that Q-learning performs better than TFT. This explains the behavior of TFT and QL when temperature decay is introduced, as described in Figure 6. In this figure, QL seems to be more effective than TFT. This is probably a result of the fact that in this experiment the student's temperature is quite low most of the time.

In these experiments the QL remembers only the last joint action. We experimented with QL with more memory and performance was worse. This can be explained as follows. For a QL with memory one or more, the problem is a fully observable Markov decision process *once* the teacher plays TFT, because TFT is a deterministic function of the previous joint action. We know that Q-learning converges to the optimal policy under such conditions (Watkins & Dayan, 1992). Adding more memory effectively adds irrelevant attributes, which, in turn, causes a slower learning rate. We have also examined whether 2TFT would be successful when agents have a memory of two. The results are not shown here, but the success rate was considerably lower than for TFT, although better than for two QLs.

TFT performed well as a teaching strategy, and we explained the motivation for using it. We now want to produce a more quantitative explanation, one that can be used to predict its success when we vary various parameters, such as the payoff matrix.

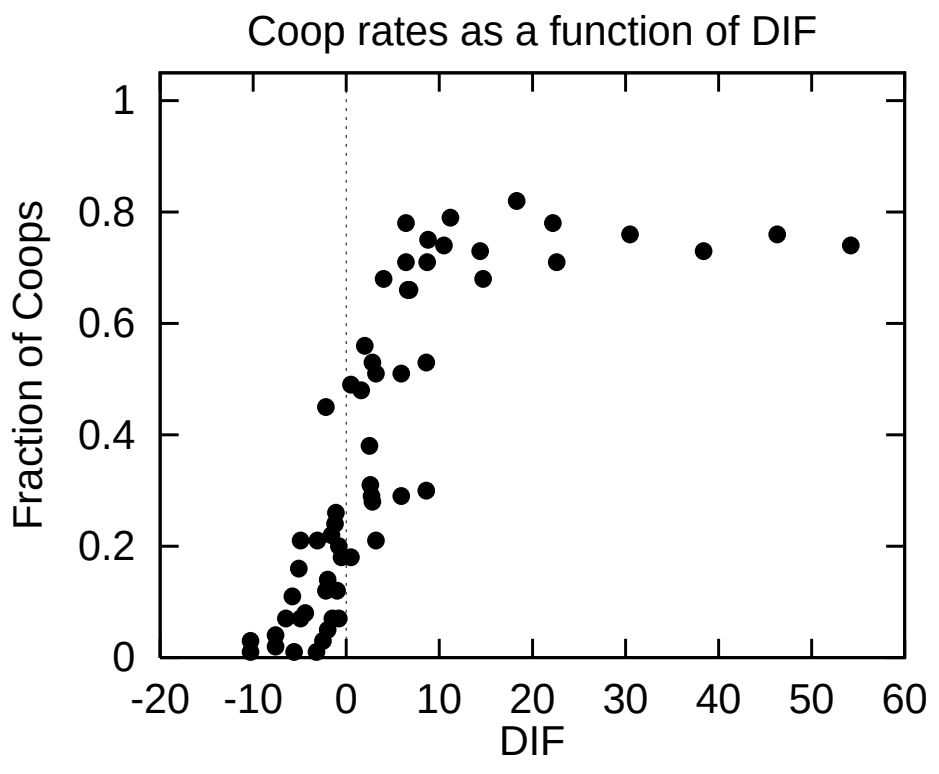

Figure 7: Coop rates as a function of DIF $= a+b+\gamma(a+c)-(c+d+\gamma(b+d))$. The means are for 100 experiments, 10000 iterations each. Student's memory is 1.

Let the student's payoff matrix be as in matrix A of Figure 1; let $p$ be the probability that the student plays Coop, and let $q = 1 - p$ be the probability that the student plays Defect. These probabilities are a function of the student's Q-values (see the description in Section 8.1). Let us assume that the probabilities $p$ and $q$ do not change considerably from one iteration to the next. This seems especially justified when the learning rate, $\alpha$, is small.

Given this information, what is the student's expected reward for playing Coop? In TFT, the teacher's current action is the student's previous action, so we can also assume that the teacher will play Coop with probability $p$. Thus, the student's expected payoff for playing Coop is $(p \cdot a + q \cdot b)$. Since Q-learners care about their discounted future reward (not just their current reward), what happens next is also important. Since we assumed that the student cooperated, the teacher will cooperate in the next iteration, and if we still assume $p$ to be the probability that the student will cooperate next, the student's expected payoff in the next step is $(p \cdot a + q \cdot c)$. If we ignore higher order $\gamma$ terms the expected reward of playing Coop becomes: $p \cdot a + q \cdot b + \gamma(p \cdot a + q \cdot c)$. The expected reward of Defect is thus: $p \cdot c + q \cdot d + \gamma(p \cdot b + q \cdot d)$. Therefore, TFT should succeed as a teaching strategy when:

$$p \cdot a + q \cdot b + \gamma(p \cdot a + q \cdot c) > p \cdot c + q \cdot d + \gamma(p \cdot b + q \cdot d).$$

Since initially $p = q = 0.5$, and it is the behavior at the stage where $p$ and $q$ are approximately equal that will determine whether TFT succeeds, we can attempt to predict the success of TFT based on whether:

$$\text{DIF} = a + b + \gamma(a + c) - [(c + d + \gamma(b + d))] \geq 0$$

To test this hypothesis we ran TFT on a number of matrices using Q-learners with different discount factors. The results in Figure 7 show the fraction of Coops over 10000 iterations as a function of DIF for a teacher using TFT, and with temperature decay. We see that DIF is a reasonable predictor of success. When it is below 0, almost all rates are below 20%, and above 8 most rates are above 65%. However, between 0 and 8 it is not successful.

### 8.4 Teaching as a Design Tool

In Section 6 we identified a class of games that are challenging to teach, and the previous sections were mostly devoted to exploring teaching strategies in these games when the student is a Q-learner. One of the assumptions we made was that the teacher is trying to optimize some function of the student's behavior and does not care what she has to do in order to achieve this optimal behavior. However, often the teacher would like to maximize some function that depends both on her behavior and on the student's behavior. When this is the case, even the more simple games discussed in Section 6 pose a challenge.

In this section, we examine a basic coordination problem, block pushing, in which our objective is not teaching, but where teaching is essential for obtaining good results. Our aim in this section is to demonstrate this point, and hence the value of understanding embedded teaching. Our results show that there is a teaching strategy that achieves much better performance than a naive teaching strategy and leads to behavior that is much better than that of two reinforcement learners.

Consider two agents that must push a block as far as possible along a given path in the course of 10,000 time units. At each time unit each agent can push the block along the path, either gently (saving its energy) or hard (spending much energy). The block will move in each iteration $c \cdot x \cdot h + (2 - x) \cdot h$ units in the desired direction, where $h, c > 0$ are constants and $x$ is the number of agents which push hard. At each iteration, the agents are paid according to the distance the block was pushed. Naturally, the agents wish to work as little as possible while being paid as much as possible, and the payoff in each iteration is a function of the cost of pushing and the payment received. We assume that each agent prefers that the block will be pushed hard by at least one of the agents (guaranteeing reasonable payment), but each agent also prefers that the other agent will be the one pushing hard. If we denote the two actions by *gentle* and *hard*, we get that the related game can be described as follows:

| | *hard* | *gentle* |
|---|---|---|
| *hard* | (3,3) | (2,6) |
| *gentle* | (6,2) | (1,1) |

Notice that the above game falls into the category of games where teaching is easy. If all the teacher cares about is that the student will learn to push hard, she will simply push gently. However, when the teacher is actually trying to maximize the distance in which the block moved, this teaching strategy may not be optimal. Notice that there can be at most 20000 instances of hard push; the naive teaching strategy mentioned above will yield no more than 10000 instances of hard push. In order to increase the number, we need a more complex teaching strategy.

In the results below we use BQL with $\alpha = 0.001$. Consider the following strategy for the teacher: push gently for $K$ iterations, and then start to push hard. As we will see, by a right selection of $K$, we obtain the desired behavior. Not only will the student push hard most of the time, but the total number of hard push instances will improve dramatically. In Figure 8, the $x$ coordinate corresponds to the parameter $K$, while the $Y$ coordinate corresponds to the number of hard push instances which occur in 10000 iterations. The results obtained are average results for 50 trials.

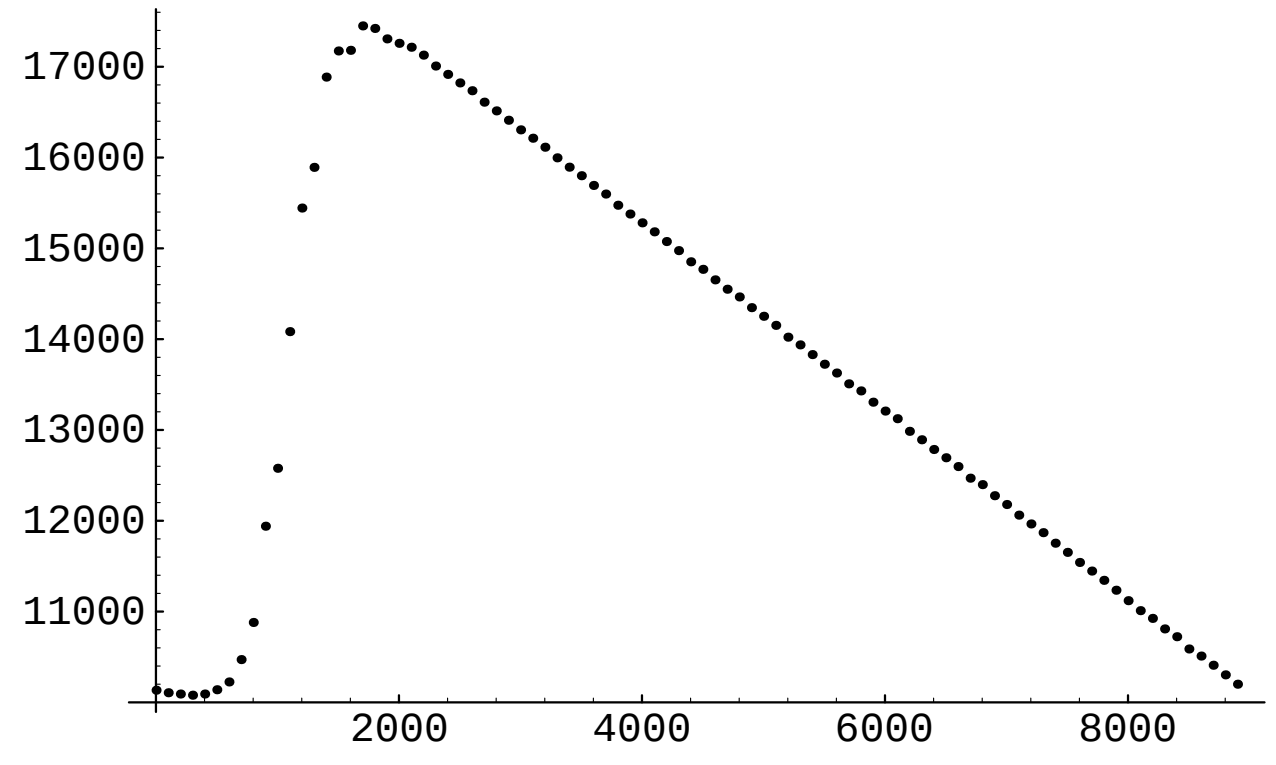

Figure 8: Teaching to push hard: number of hard push instances by the student in 10000 iterations as a function if the number of iterations in which the teacher does not push hard (avg. over 50 trials).

As we can see from Figure 8, the efficiency of the system is non-monotonic in the threshold $K$. The behavior we obtain with an appropriate selection of $K$ is much better than what we would have obtained with the naive teaching strategy. It is interesting to note the existence of a sharp phase transition in the performance at the neighborhood of the optimal $K$. Finally, we mention that when both agents are reinforcement learners, we get only 7618 instances of "push hard", which is much worse than what is obtained when we have a knowledgeable agent that utilizes its knowledge to influence the behavior of the other agent.

## 9. Towards A General Theory

The two case studies presented in this paper raise the natural question of whether general, domain independent techniques for PCMAS design exist, and whether we have learned about such tools from our case studies. We believe that it is still premature to say whether a general theory of PCMAS design will emerge; this requires much additional work. Indeed, given the considerable differences that exist between the two domains explored in this paper, and given the large range of multi-agent systems and agents that can be envisioned, we doubt the existence of common low-level techniques for PCMAS design. Even within the class of rational agents which we investigated, agents can differ considerably in their physical, computational, and memory capabilities, and in their approach to decision making (e.g., expected utility maximization, maximization of worst-case outcomes, minimization of regret). Similarly, the problem of social-law enforcement can take on different forms, for example, the malicious agents could cooperate among each other. However, once a more abstract view is taken, certain important unifying concepts appear, namely, punishment and reward.

Punishment and reward are abstract descriptions of two types of high-level feedbacks that the controllable agents can provide to the uncontrollable agents. Although punishment and reward take different form and meaning in the two domains, in both cases, the uncon-trollable agents seem to "care" about the controllable agent's reaction to their action. What we see is that in both cases, the controllable agents can influence the uncontrollable agents' perception of the worthiness of their actions. The precise manner in which the controllable agents affect this perception differs, but in both cases it utilizes some inherent aspect of uncertainty in the uncontrollable agent's world model. In the case of rational agents, despite their perfect knowledge of the dynamics of the world, uncertainty remains regarding the outcome of the non-malicious agents' actions. By fixing a certain course of action for the controllable agents, we influence the malicious agents' perception of the outcome of their own actions. In the case of the learning agent, one can affect the perception of the student's action by affecting its basic world model. Hence, it seems that a high-level approach for PCMAS design has two stages: First, we analyze the factors that influence the uncontrollable agent's perception of their actions. Next, we analyze our ability to control these factors. In retrospect, this has been implicit in our approach. In our study of social-law enforcement, we used the projected game to find out how an agent's perception of an action can be changed and used the indirect mechanism of threats to enforce the perception we desired. In our study of embedded teaching, we started with an analysis of different games and the possibility of affecting an agent's perception of an action in these games. Next, we tried to provide this perception. In the case of BQL students, our controllable teacher did not have complete control over the elements that determine the student's perception because of the random nature of the student's action. Yet, she did try to somehow affect them. In the case of the Q-learners, direct control was not available over all factors determining the student's perception. Yet, the teacher could control some aspects of this perception, which were found to be sufficient.

One might ask how representative our studies are of general PCMAS domains, and therefore, how relevant is the insight they may provide. We have chosen these two domains with the belief that they represent key aspects of the types of agents studied in AI. In AI, we study dynamic agents that act to improve their state. These agents are likely to use information to revise their assessment of the state of the world, much like the learning agents, and will need to make decisions based on their current information, much like the expected utility maximizers we have studied. Hence, typical multi-agent systems studied in AI include agents that exhibit one or both of these properties.

While punishment and rewards provide the conceptual basis for designing the controllable agents, MDPs supply a natural model for many domains. In particular, MDPs are suitable when uncertainty exists, stemming either from the other agents' choices or from nature. As we showed in Section 7, at least in principle, we can use established techniques to obtain strategies for the controllable agents when the problem can be phrased as a Markov decision process. Using the MDP perspective in other cases would require more sophisticates tools and a number of important challenges must be met first: (1) The assumptions that the agent's state is fully observable and that the environment's state is fully observable is unrealistic in many domains. When these assumptions are invalid, we obtain a partially observable Markov decision process (POMDP) (Sondik, 1978). Unfortunately, although POMDPs can be used in principle to obtain the ideal policy for our agents, current techniques for solving POMDPs are limited to very small problems. Hence, in practice one will have to resort to heuristic punishment and reward strategies. (2) In Section 7 we had only one controlling agent. This poses a natural challenge of generalizing tools and techniques from MDPs to distributed decision making processes.

## 10. Summary and Related Work

This paper introduces the distinction between controllable and uncontrollable agents and the concept of partially controlled multi-agent systems. It provides two problems in multi-agent system design that naturally fall into the framework of PCMAS design and suggests concrete techniques for influencing the behavior of the uncontrollable agents in these domains. This work contributes to AI research by introducing and exploring a promising perspective on system design and it contributes to DES research by considering two types of structural assumptions on agents, corresponding to rational and learning agents.

The application of our approach to the enforcement of social behavior introduces a new tool in the design of multi-agent systems, punishment and threats. We used this notion and investigated it as part of an explicit design paradigm. Punishment, deterrence, and threats have been studied in political science (Dixit & Nalebuff, 1991; Schelling, 1980); yet, in difference to that line of work (and its related game-theoretic models), we consider the case of a dynamic multi-agent system and concentrate on punishment design issues, such as the question of minimizing the number of reliable agents needed to control the system. Unlike much work in multi-agent systems, we did not assume all agents to be rational or all agents to be law-abiding. Rather, we only assumed that the designer can control some of the agents and that deviations from the social laws by the uncontrolled agents need to be rational. Notice that the behavior of controllable agents may be considered irrational in some cases; however, it will eventually lead to desired behavior for all the agents. Some approaches to negotiations can be viewed as incorporating threats. In particular, Rosenschein and Genesereth (1985) consider a mechanism making deals among rational agents, where agents are asked to offer a joint strategy to be followed by all agents and declare the move they would take if there will be no agreement on the joint strategy. This latter move can be viewed as a threat describing the implications of refusing the agent's suggested joint strategy. For example, in the prisoner's dilemma setting an agent may propose joint cooperation and threaten defecting otherwise. The work in the first part of this paper could be viewed as examining how such a threat could be credible and effective in a particular context of iterative multi-agent interactions.

As part of our study, we proposed embedded teaching as a situated teaching paradigm suitable for modeling a wide range of teaching instances. We modeled the teacher and the student as players in an iterated two-player game. We concentrated on a particular iterative game, which we showed to be the most challenging game of its type. In our model, the dynamics of the teacher-student interaction is made explicit, and it clearly delineated the limits placed on the teacher's ability to influence the student. We showed that with a detailed model of the student, optimal teaching policies can be theoretically generated by viewing the teaching problem as a Markov decision process. The performance of the optimal teaching policy serves as a bound on any agent's ability to influence this student. We examined our ability to teach two types of reinforcement learners. In particular, we showed that when an optimal policy cannot be used, we can use TFT as a teaching method. In the case of Q-learners this policy was very successful. Consequently, we proposed a model that explains this success. Finally, we showed that even in those games in which teaching is not challenging, it is nevertheless quite useful. Moreover, when our objective is more than simply teaching the student, even those simpler domains present some non-trivial choices. In the future we hope to examine other learning architectures and see whether the lessons learned in this domain can be generalized, and whether we can use these methods to accelerate learning in other domains.

A number of authors have discussed reinforcement learning in multi-agent systems. Yanco and Stein (1993) examine the evolution of communication among cooperative reinforcement learners. Sen *et al.* (1994) use Q-learning to induce cooperation between two block pushing robots. Matraic (1995) and Parker (1993) consider the use of reinforcement learning in physical robots. They consider features of real robots, which are not discussed in this paper. Shoham and Tennenholtz (1992) examine the evolution of conventions in a society of reinforcement learners. Kittock (1994) investigates the effects of societal structure on multi-agent learning. Littman (1994) develops reinforcement learning techniques for agents whose goals are opposed, and Tan (1993) examines the benefit of sharing information among reinforcement learners. Finally, Whitehead (1991) has shown that $n$ reinforcement learners that can observe everything about each other can decrease learning time by a factor of $n$. However, the above work is not concerned with teaching, or with the question of how much influence one agent can have over another. Lin (1992) is explicitly concerned with teaching as a way of accelerating learning of enhanced Q-learners. He uses experience replay and supplies students with examples of how the task can be achieved. As we remarked earlier, this teaching approach is different from ours, since the teachers are not embedded in the student's domain. Within game theory there is an extensive body of work that tries to understand the evolution of cooperation in the iterated prisoner's dilemma and to find good playing strategies for it (Eatwell et al., 1989). In that work both players have the same knowledge, and teaching is not an issue.

Last but not least, our work has important links to work on conditioning and especially operant conditioning in psychology (Mackintosh, 1983). In conditioning experiments an experimenter tries to induce changes in its subjects by arranging that certain relationships will hold in their environment, or by explicitly (in operant conditioning) reinforcing the subjects' actions. In our framework the controlled agent plays a similar role to that of the experimenter. Our work uses a control-theoretic approach to the related problem, while applying it to two basic AI contexts.

The main drawback of our case studies is the simple domains in which they were conducted. While this is typical of initial exploration of new problems, future work should try to remove some of the limiting assumptions that our models incorporate. For example, in the embedded teaching context, we assumed that there is no uncertainty about the outcome of a joint action. Similarly, our model of multi-agent interaction in Section 3 is very symmetric, assuming all agents can play all of the $k$ roles in the game, that they are equally likely to play each role, etc. Another assumption made was that malicious agents were "loners" acting on their own, as opposed to a team of agents. Perhaps more importantly, future work should identify additional domains that are naturally described in terms of PCMAS and formalize a general methodology for solving PCMAS design problems.

## Acknowledgements

We are grateful to Yoav Shoham and other members of the Nobotics group at Stanford for their input, and to the anonymous referees for their productive comments and suggestions. We are especially grateful to James Kittock for his comments and his help in improving the presentation of this paper. This research was supported by the fund for the promotion of research at the Technion, by NSF grant IRI-9220645, and by AFOSR grant AF F49620-94-1-0090.